\documentclass[sigconf, natbib=true]{acmart}

\AtBeginDocument{%
  \providecommand\BibTeX{{%
    \normalfont B\kern-0.5em{\scshape i\kern-0.25em b}\kern-0.8em\TeX}}}

\usepackage{xcolor}         % colors
\usepackage{epsfig}
\usepackage{graphicx}
\usepackage{amsfonts}       % blackboard math symbols
\usepackage{amsmath}
\usepackage{multirow}
\usepackage{caption}

\usepackage{subcaption}

\newcommand{\etal}{\emph{et al.}}
\newcommand{\eg}{\emph{e.g.}}
\newcommand{\ie}{\emph{i.e.}}
\setcopyright{iw3c2w3}
\copyrightyear{2023}
\acmYear{2023}
\acmDOI{XXXXXXX.XXXXXXX}

\acmConference[Preprint]{Arxiv}{2023}{Arxiv}
\acmPrice{15.00}
\acmISBN{978-1-4503-XXXX-X/18/06}

\acmSubmissionID{??}

\begin{document}

\title{Cascaded Cross-Modal Transformer for Request and Complaint Detection}

\author{Nicolae-C\u{a}t\u{a}lin Ristea}
\affiliation{%
  \institution{Politehnica University of Bucharest,\\ University of Bucharest}
  \country{Romania}
}

\author{Radu Tudor Ionescu}
\email{raducu.ionescu@gmail.com}
\affiliation{%
  \institution{University of Bucharest}
  \country{Romania}
}

\begin{abstract}
%Customer relationship management is mandatory for a company's success, as it facilitates effective communication between the organization and its customers, thereby fostering stronger relationships. Call centers serve as a primary platform for such interactions, with phone calls being a common mode of communication. Consequently, speech analysis applications have emerged as powerful tools for comprehending customer interactions, understanding user interests, and identifying potential issues. To address these needs, 
We propose a novel cascaded cross-modal transformer (CCMT) that combines speech and text transcripts to detect customer requests and complaints in phone conversations. Our approach leverages a multimodal paradigm by transcribing the speech using automatic speech recognition (ASR) models and translating the transcripts into different languages. Subsequently, we combine language-specific BERT-based models with Wav2Vec2.0 audio features in a novel cascaded cross-attention transformer model. We apply our system to the Requests Sub-Challenge of the ACM Multimedia 2023 Computational Paralinguistics Challenge, reaching unweighted average recalls (UAR) of $65.41\%$ and $85.87\%$ for the complaint and request classes, respectively. %Our code is freely available at: \url{http//github.com/link.hidden.for.review}.
\end{abstract}




\maketitle

\section{Introduction}
In recent years, the field of computational paralinguistics has witnessed significant advancements in analyzing and interpreting non-verbal vocal cues, leading to valuable insights into human communication. As part of this research landscape, we present a multimodal framework for the Requests Sub-Challenge (RSC) of the ACM Multimedia 2023 Computational Paralinguistics Challenge (ComParE) \cite{Schuller-ACMMM-2023}. In this sub-challenge, the task is to detect the presence or absence of a request or complaint within audio calls between agents and customers. The objective is to develop an effective model that can accurately identify and categorize instances where a customer expresses a request or complaint during the course of the conversation.

Being inspired by the success of previous multimodal methodologies \cite{Yoon-SLT-2018, Akbari-NeurIPS-2021, Das-ACM-2023, Jabeen-TACM-2023,Georgescu-arXiv-2022} on other tasks, we propose a novel multimodal framework which effectively harnesses cross-domain features derived from both speech and text data, which are subsequently integrated into a cascaded cross-modal transformer (CCMT) model. To obtain multimodal information from audio data, the only modality provided by the RSC organizers, we employ state-of-the-art automatic speech recognition (ASR) models \cite{Baevski-NeurIPS-2020, Radford-ARXIV-2022} to transcribe the provided audio conversations. The additional modality, obtained through speech-to-text conversion, provides valuable insights that complement the original audio data, enabling the application of various natural language processing (NLP) techniques. Furthermore, recognizing the existence of large language models (LLMs) tailored for distinct languages \cite{Martin-ACL-2020, Devlin-NAACL-2019, Canete-ICLR-2020}, we expand the scope of our research by translating the transcripts into multiple languages, such as English and Spanish, via neural machine translation (NMT). %To accomplish this, we utilize the FLAN T5 language model \cite{Chung-ARXIV-2022}, enabling us to effectively analyze the same conversation in different linguistic contexts.

Tackling the complexity of real-world data through the combination of multiple modalities is a challenging task, requiring the development of a robust and efficient method for aggregating all sources of information \cite{Ramachandram-SPM-2017, Gao-NC-2020, Stahlschmidt-BB-2022}. To address this challenge, we propose a novel CCMT model that aggregates information from two NLP models, namely CamemBERT \cite{Martin-ACL-2020} and BERT \cite{Devlin-NAACL-2019}, in the first cascade step, and further combines the resulting multi-language textual features with the audio-based Wav2Vec2.0 \cite{Baevski-NeurIPS-2020} features in the second cascade step. %By integrating audio-text modalities, our model captures the rich semantic and acoustic characteristics of the data, enabling a comprehensive understanding of the customer-agent interactions. 
While the employed NLP models facilitate capturing nuanced language cues and contextual information within the conversations, the Wav2Vec2.0 model complements the textual data by providing insights into vocal tone, emphasis, and other non-verbal cues that contribute to the overall sentiment and intent expressed by the customers.

% \begin{figure}[!t]
% \begin{center}
% \centerline{\includegraphics[width=1.0\linewidth]{fig_LeRaC.pdf}}
% %\vspace{-0.25cm}
% \caption{Training based on Learning Rate Curriculum.}
% \label{fig_lerac}
% %\vspace{-1.1cm}
% %\vspace{-0.9cm}
% \end{center}
% \end{figure}

In summary, our contribution is threefold:
\begin{itemize}
    \item We propose a novel framework that generates multiple text modalities from audio via ASR and NMT, enabling us to leverage different linguistic contexts for the 2023 Computational Paralinguistics Challenge \cite{Schuller-ACMMM-2023}.
    \item We introduce a novel cross-modal transformer architecture, called CCMT, which aggregates text and audio through a cascaded cross-attention mechanism.
    \item We provide strong empirical evidence in favor of our framework, via a comprehensive set of experiments. 
\end{itemize}

\begin{figure*}[!t]
\begin{center}
\centerline{\includegraphics[width=1.0\linewidth]{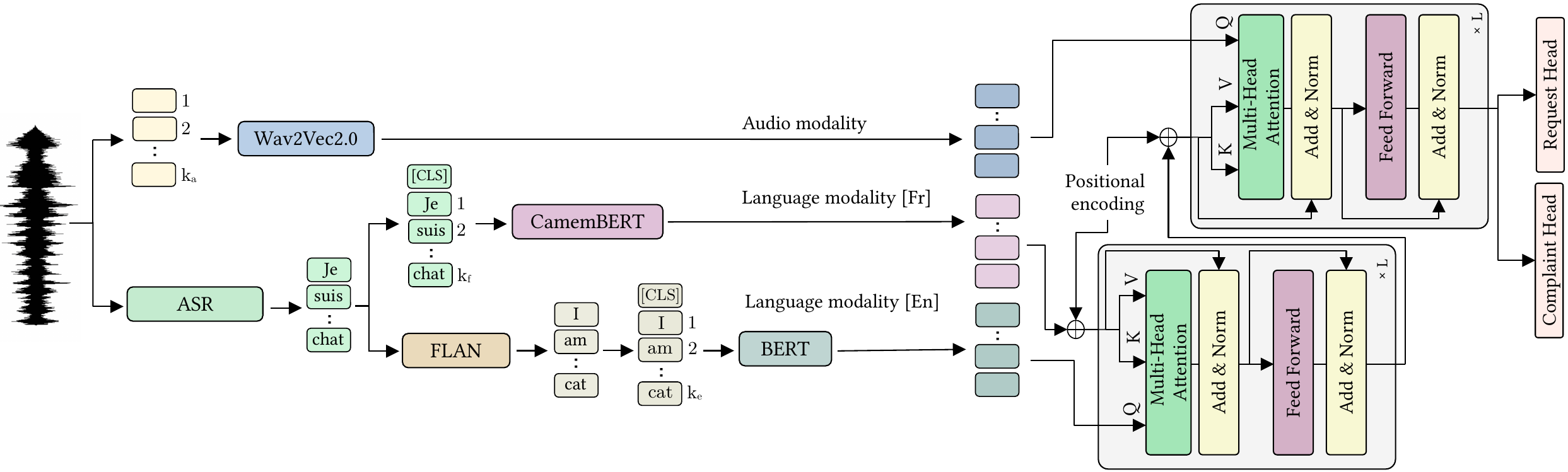}}
\vspace{-0.3cm}
\caption{Our multimodal framework for request and complaint classification. For the audio modality, we extract tokens using the Wav2Vec2.0 \cite{Baevski-NeurIPS-2020} model on time-domain audio input data. For the text modalities, we first apply an ASR model to transcribe each audio into French text. For the French language modality [Fr], the tokens are directly processed by the CamemBERT \cite{Martin-ACL-2020} model. For the English language modality [En], we utilize a language translation model called FLAN \cite{Chung-ARXIV-2022} to translate the French text into English. The English text tokens are then fed into the BERT \cite{Devlin-NAACL-2019} model. The resulting tokens are processed by the CCMT model, which feeds the final class token into the complaint and request classification heads.} 
\label{fig_framework}
%\vspace{-1.1cm}
\vspace{-0.5cm}
\end{center}
\end{figure*}

\vspace{-0.15cm}
\section{Method}

We design a novel multimodal framework for request and complaint classification, which is illustrated in Figure \ref{fig_framework}. Starting from the audio input data, our framework derives two additional text modalities via automatic speech recognition (ASR) and neural machine translation (NMT). The three modalities are further processed by our cascaded cross-modal transformer (CCMT) model. Next, we describe each component in more detail.

\noindent
{\bf Audio branch.}
In our framework, we employ the Wav2Vec2.0 \cite{Baevski-NeurIPS-2020} model to learn representative tokens for the audio modality. The raw audio data is split into $k_a \in \mathbb{N^+}$ chunks, where $k_a$  depends on the input length and varies from one sample to another. The initial tokens are fed into the model, which further performs a series of computations to extract meaningful audio representations. %This process involves several stages, including preliminary convolutional layers for local feature extraction, subsequent transformer layers for capturing global contextual information, and a quantization step for efficient representation. 
The output of the Wav2Vec2.0 model consists of the same number of $k_a$ tokens, representing the acoustic features of the audio modality. These tokens encode important information about the audio signals, such as pitch, frequency, and intensity. We hereby note that encoders of distinct modalities provide different numbers of output tokens. However, our CCMT model requires the same number of tokens for each modality. We randomly sample a fixed number of $k \in \mathbb{N^+}$ tokens to comply with the data uniformity constraint enforced by CCMT. The selected audio tokens are then fed into the CCMT model. %The tokens are further combined with tokens from the text modalities to facilitate the multimodal analysis. By fine-tuning the Wav2Vec2.0 model, our framework benefits from a rich representation of the audio modality, enabling the CCMT model to effectively capture and integrate both linguistic and acoustic information.

\noindent
{\bf Text branches.}
To extract text transcripts from the audio files, we employ a series of ASR models based on the Whisper architecture \cite{Radford-ARXIV-2022}, with three different backbones: small, medium and large. These models generate French transcripts, since the language spoken in the audio files is French. We consider multiple ASR models as an augmentation technique to enhance the training data.
Subsequently, we incorporate two language modalities: French and English. Spanish is also considered as an option, but we do not integrate it in the final model because it lowers the overall performance. For the French branch, we have a total of $k_f + 1 \in \mathbb{N^+}$ tokens, consisting of $k_f$ words and an additional class token. These tokens are given as input to the CamemBERT model \cite{Martin-ACL-2020}, resulting in $k_f$ output tokens. To ensure uniformity across modalities, we randomly sample a number of $k$ tokens from the output, thus obtaining the same number of final tokens as for the audio modality.

For the English text modality, we employ the FLAN T5 \cite{Chung-ARXIV-2022} language model to translate French text into English. This process can naturally result in a different number of words. Consequently, the input for the BERT model \cite{Devlin-NAACL-2019} consists of $k_e + 1 \in \mathbb{N^+}$ tokens, comprising $k_e$ words and one class token. As for the French language modality, we sample a fixed number of $k$ tokens from the output to maintain uniformity across modalities. If the number of tokens for either French or English modality is less than $k$, we randomly duplicate tokens until $k_f = k_e = k$ to meet the uniformity constraint. By incorporating both French and English language modalities, we ensure that the CCMT model can effectively capture and integrate linguistic information from multiple languages, facilitating a comprehensive multimodal analysis for request and complaint classification.

% \begin{figure}[!t]
% \begin{center}
% \centerline{\includegraphics[width=0.8\linewidth]{ccrossmt.pdf}}
% \vspace{-0.3cm}
% \caption{As input, the CCMT architecture receives tokens obtained from CamemBERT, BERT, and Wav2Vec2.0. To maintain the positional information of each modality, we introduce separate positional embeddings. The tokens are processed by two cascaded cross-attention transformer blocks. The first block combines the French and English text modalities, and the resulting tokens are combined with the audio modality. The final class token is passed to the two MLP classification heads to make the final predictions.}
% \label{fig_ccmt}
% %\vspace{-1.1cm}
% \vspace{-0.5cm}
% \end{center}
% \end{figure}

\noindent
{\bf Cascaded Cross-Modal Transformer.}
Given that all models generate tokens with the same dimensionality, let $T_f \in \mathbb{R}^{k \times d}$ represent the set of tokens generated by the CamemBERT model, $T_e \in \mathbb{R}^{k \times d}$ represent the set of tokens generated by the BERT model, and $T_a \in \mathbb{R}^{k \times d}$ represent the set of audio tokens generated by the Wav2Vec2.0 model, where $d \in \mathbb{N^+}$. To let our model distinguish between token positions from different modalities, we introduce positional encoding vectors that are distinct among modalities.

In the first transformer block, we introduce the learnable parameters $W_Q', W_K', W_V' \in \mathbb{R}^{d \times d_h}$ for the projection blocks, where $d_h \in \mathbb{N^+}$ represents the dimension of a single attention head. To obtain the query, keys, and values, we perform matrix multiplications between the input tokens and the projection matrices: $Q' = T_e \cdot W_Q'$, $K' = T_f \cdot W_K'$, $V' = T_f \cdot W_V'$. We use the English modality for queries, and the French modality for keys and values, as we consider that the French modality is more important for the task, precisely because the input phone calls are in French. The output of the cross-attention layer is denoted as $U' \in \mathbb{R}^{k \times d_h}$ and can be expressed as follows:
\begin{equation}
\label{eq1}
U' = \mbox{softmax}\left( \frac{Q'\cdot K'^{\top}}{\sqrt{d_h}} \right) \cdot V'.
\end{equation}

To ensure that the tokens maintain the same output dimensionality, we introduce a learnable matrix $M' \in \mathbb{R}^{d_h \times d}$ within the multi-head attention layer. By multiplying the output $U'$ with $M'$, we restore the original dimensionality of the input tokens, resulting in $Y' = U' \cdot M'$. Subsequently, we apply the summation and normalization operations, followed by a feed-forward module (FF), and another summation and normalization layer. The equations that describe these operations are formally presented below:
\begin{equation}
\label{eq2}
Z' = Y' + \mbox{Norm}(Y'),\; T_c = Z' + \mbox{FF}(\mbox{Norm}(Z')),
\end{equation}
where $T_c \in \mathbb{R}^{k \times d}$ denotes the output linguistic cross-attention tokens. In the second cross-attention transformer block, these tokens are combined with the $T_a$ tokens. Here, we introduce the learnable parameters $W_Q'', W_K'', W_V'' \in \mathbb{R}^{d \times d_h}$ for the projection blocks. As for the first transformer block, we obtain the query, keys, and values by multiplying the input tokens with the corresponding projection matrices: $Q'' = T_a \cdot W_Q''$, $K''= T_c \cdot W_K''$, $V'' = T_a \cdot W_V''$. By employing operations analogous to Equations \eqref{eq1} and \eqref{eq2}, we obtain the output tokens $T_o \in \mathbb{R}^{k \times d}$. The first token, which represents the class token, is passed to two multi-layer perceptron (MLP) heads. The MLP heads produce the final prediction for the request and complaint classes. % The operations are presented in the context of a single attention head, the extension to multiple heads being trivial.

\vspace{-0.15cm}
\section{Experiments}
\label{sec_experiments}
%\vspace{-0.05cm}
%\subsection{Experimental Setup}
%\vspace{-0.1cm}

\noindent
{\bf Data set.}
The data set provided by the ComParE organizers for RSC is a subset (audio-only) of the HealthCall30 corpus, constructed by Lackovic \etal~\cite{Lackovic-arXiv-2022}. The data set is partitioned into a training set of 6,822 samples, a development set of 3,084 samples and a test  set of 3,503 samples. % It comprises real audio recordings between call center agents and customers who call to solve a problem or to request information. Each data sample is a recording of $30$ seconds at a sampling rate of $16$ kHz. Each conversation has two separate audio channels. The first channel corresponds to the customer’s audio and the second corresponds to the agent’s audio.

\noindent
{\bf Performance measure.} The participants are ranked by the organizers based on the unweighted average recall (UAR), \ie~the average of the recall scores on the positive and negative classes. % We therefore report our performance in terms of this measure.

\noindent
{\bf Baselines.}
% In our study, we evaluate the proposed method against the leading baseline results provided by the ComParE organizers \cite{Schuller-ACMMM-2023}. 
For the audio experiments, we compare our model, which is based on Wav2Vec2.0 \cite{Baevski-NeurIPS-2020}, against ResNet-50 \cite{He-CVPR-2016} and various transformer-based approaches \cite{Gong-INTERSPEECH-2021, Ristea-INTERSPEECH-2022}. For the text experiments, we conduct a comparative analysis of multiple NLP models \cite{Le-LREC-2020, Martin-ACL-2020} and four ASR models \cite{Baevski-NeurIPS-2020, Radford-ARXIV-2022}. We also explore various fusion techniques,  comparing our CCMT model with a conventional transformer model \cite{Dosovitskiy-ICLR-2020}, and other common fusion techniques, \eg~based on plurality voting or multi-layer perceptrons (MLPs).

% \subsubsection{Data preprocessing}
% For the audio modality, we take the 16 kHz audio file comprising both agent and customer speeches, which are overlapped into a single channel. The single channel is obtained by averaging the original audio channels. The other preprocessing steps are identical to those of Baevski \etal~\cite{Baevski-NeurIPS-2020}. For all ASR models \cite{Baevski-NeurIPS-2020, Radford-ARXIV-2022}, we consider the audio samples that are split into two separate channels, one for the agent and one for the customer. Afterwards, we resample each audio channel at 16 kHz and feed it to the ASR models, following the preprocessing steps described in the original papers \cite{Baevski-NeurIPS-2020, Radford-ARXIV-2022}. For the text models, we %tried out multiple ways of feeding the text data, but the best results were obtained by 
% simply concatenate the transcripts of the agent and the customer. 

% \subsubsection{Data augmentation}
% For the audio features, we employ Gaussian noise, clipping, pitch shifting, low-pass filtering, high-pass filtering and volume change as augmentation methods. For the text modalities, we use classical augmentation methods, \eg~insertion, deletion, word swapping, in addition to an augmentation based on varying the ASR model. More precisely, we use randomly picked transcripts from multiple ASR models. This augmentation technique is also used for the NLP models trained on other languages, simply by translating the transcripts from multiple ASR models.

\noindent
{\bf Hyperparameter choices.}
The Wav2Vec2.0 \cite{Baevski-NeurIPS-2020} model is fine-tuned for $10$ epochs with a learning rate of $10^{-5}$ on mini-batches of $16$ samples. The BERT \cite{Devlin-NAACL-2019} and CamemBERT \cite{Martin-ACL-2020} models are both trained for $25$ epochs with a learning rate of $5\cdot10^{-5}$ and a weight decay of $10^{-5}$, on mini-batches of $32$ examples. For all other models, we use the hyperparameters recommended by the authors introducing the respective models. The CCMT model is trained for $30$ epochs with a learning rate of $10^{-4}$ on mini-batches of $32$ samples. All models are trained with the Adam optimizer \cite{Kingma-ICLR-2014}. For the CCMT model, we randomly sample $k=100$ tokens (always keeping the class token) for each input modality.

\begin{table}[!t]
  \caption{Results on the development set with several architectures based on the audio modality. The architectures are either trained from scratch or fine-tuned. The models marked with an asterisk ($*$) are pretrained on other data sets. We report the mean UAR (in percentages) and the standard deviation over three runs. The best score on each task is highlighted in bold.\vspace{-0.3cm}}
  \label{tab_audio}
\small{
  \begin{center}
  \setlength\tabcolsep{3.5pt}
  \begin{tabular}{|l|l|c|c|}
    \hline
    \multirow{2}{*}{Model}       & \multirow{2}{*}{Input data}    & \multicolumn{2}{c|}{UAR} \\
    \cline{3-4}
    & & Request & Complaint\\
    \hline    
    \hline
    ResNet-50          & Spectrogram      &  $59.51 \pm 1.27$ & $52.18 \pm 0.86$ \\
    ResNet-50          & STFT      &  $60.84 \pm 1.08$   & $53.49 \pm 0.73$ \\
    ResNet-50          & Mel-Spectrogram    &  $60.31 \pm 1.01$  & $53.44 \pm 0.74$ \\
    SepTr \cite{Ristea-INTERSPEECH-2022}      & STFT     &  $62.31 \pm 0.59$   & $54.03 \pm 0.55$ \\
    AST \cite{Gong-INTERSPEECH-2021}$^*$    & Spectrogram    &  $64.72 \pm 0.45$   & $55.91 \pm 0.39$ \\
    1D Transformer & Time domain    &  $ 61.63 \pm 0.42$   &  $53.82 \pm 0.39$ \\
    Wav2Vec2.0 \cite{Baevski-NeurIPS-2020} & Time domain   &  $ 68.87 \pm 0.21$   & $ 56.55 \pm 0.22$ \\
    Wav2Vec2.0 \cite{Baevski-NeurIPS-2020}$^*$  & Time domain    & $ \textbf{71.64} \pm 0.16$   & $ \textbf{58.12} \pm 0.16$ \\
    \hline
  \end{tabular}
  \end{center}
}
\vspace{-0.2cm}
\end{table}

% Results for audio modality experiments with several architectures. We trained from scratch, as well as fine-tuned models. We reported the UAR with the standard deviation between three runs.

\begin{table}[!t]
  \caption{Results on the development set with several NLP models trained on French transcripts generated with Wav2Vec2.0 \cite{Baevski-NeurIPS-2020} and Whisper \cite{Radford-ARXIV-2022} ASR models. Whisper S+M+L stands for our augmentation technique based on using the transcripts from all three ASR models. We report the mean UAR (in percentages) and the standard deviation over three runs. The best score on each task is highlighted in bold.\vspace{-0.3cm}}
  \label{tab_text}
  \setlength\tabcolsep{3.5pt}
  \small{
  \begin{center}
  \begin{tabular}{|l|l|c|c|}
    \hline
    \multirow{2}{*}{Model}       & \multirow{2}{*}{ASR model}    & \multicolumn{2}{c|}{UAR} \\
    \cline{3-4}
    & & Request & Complaint\\
    
    % Model       & ASR Model & Request UAR & Complaint UAR\\
    \hline    
    \hline
    LSTM  \cite{Hochreiter-NC-1997}               & Wav2Vec2.0 \cite{Baevski-NeurIPS-2020}       &    $71.14 \pm 0.51$   & $55.49 \pm 0.50$ \\
    FlauBERT \cite{Le-LREC-2020}           & Wav2Vec2.0 \cite{Baevski-NeurIPS-2020}       &    $76.82 \pm 0.21$   & $58.77 \pm 0.23$ \\
    CamemBERT \cite{Martin-ACL-2020}          & Wav2Vec2.0 \cite{Baevski-NeurIPS-2020}       &    $77.45 \pm 0.13$   & $60.15 \pm 0.11$ \\
    \hline
    CamemBERT \cite{Martin-ACL-2020}          & Whisper S \cite{Radford-ARXIV-2022}        &    $79.71 \pm 0.19$   & $62.92 \pm 0.20$ \\
    CamemBERT  \cite{Martin-ACL-2020}         & Whisper M \cite{Radford-ARXIV-2022}        &    $81.86 \pm 0.11$   & $64.83 \pm 0.11$ \\
    CamemBERT \cite{Martin-ACL-2020}          & Whisper L  \cite{Radford-ARXIV-2022}        &    $82.03 \pm 0.10$   & $65.47 \pm 0.09$ \\
    CamemBERT  \cite{Martin-ACL-2020}         & Whisper S+M+L     &    $\textbf{82.44} \pm 0.08$   & $\textbf{65.61} \pm 0.08$ \\
    \hline
  \end{tabular}
  \end{center}
}
\vspace{-0.2cm}
\end{table}

%\vspace{-0.2cm}
%\subsection{Results}

\noindent
{\bf Results for the audio modality.}
The results of the models based on the audio modality are summarized in Table \ref{tab_audio}. Among the evaluated architectures, the transformer-based models demonstrate consistently better performance. Specifically, the pretrained AST \cite{Gong-INTERSPEECH-2021} model achieves a request UAR of $64.72\%$ and a complaint UAR of $55.91\%$, outperforming both SepTr \cite{Ristea-INTERSPEECH-2022} and 1D transformer models. However, the best results are obtained with the Wav2Vec2.0 \cite{Baevski-NeurIPS-2020} model. By fine-tuning the Wav2Vec2.0 model, we achieve a request UAR of $71.64\%$ and a complaint UAR of $58.12\%$, showcasing the effectiveness of the time-domain audio representation provided by Wav2Vec2.0. %These results underline the benefits of using pretrained models in audio classification tasks. 
Based on the results reported in Table \ref{tab_audio}, we select the fine-tuned Wav2Vec2.0 model for our multimodal pipeline.

\noindent
{\bf Results for the French text modality.}
In Table~\ref{tab_text}, we present the results for the French text transcripts using various NLP models. We explored different ASR models, including Wav2Vec2.0 \cite{Baevski-NeurIPS-2020} and three sizes of Whisper \cite{Radford-ARXIV-2022} (small, medium, and large). %When using the Wav2Vec2.0 ASR model, the CamemBERT model \cite{Martin-ACL-2020} achieves the best performance, with a Request UAR of $77.45\%$ and a Complaint UAR of $60.15\%$. 
Since CamemBERT achieves the best performance, we choose the CamemBERT model in favor of the FlauBERT and LSTM models For the subsequent experiments.
The Whisper family of ASR models leads to significant performance improvements. The highest results are obtained by jointly using the transcripts generated by all Whisper models.  We highlight that the results obtained for the text modality in Table \ref{tab_text} are significantly higher compared to the audio modality results in Table \ref{tab_audio}. This observation suggests that, for our specific tasks, language features are more important than acoustic features.

\begin{table}[!t]
  \caption{Results on the development set with various NLP models on three distinct languages: English (En), French (Fr) and Spanish (Sp). We report the fusion results of the French model with the other language models via an MLP-based aggregation method. We report the mean UAR (in percentages) and the standard deviation over three runs. The best score on each task is highlighted in bold.\vspace{-0.3cm}}
  \label{tab_language}
\small{
  \begin{center}
  \setlength\tabcolsep{2.5pt}
  \begin{tabular}{|l|c|c|c|}
    \hline
    \multirow{2}{*}{Model}       & \multirow{2}{*}{Language}    & \multicolumn{2}{c|}{UAR} \\
    \cline{3-4}
    & & Request & Complaint\\
    % Model       & Language & Request UAR & Complaint UAR\\
    \hline    
    \hline
    CamemBERT \cite{Martin-ACL-2020}   & Fr  &  $82.44\!\pm\!0.08$   & $65.61\!\pm\!0.08$  \\
    RoBERTa \cite{Liu-ARXIV-2019}            & En    &    $78.57\!\pm\!0.07$   & $63.89\!\pm\!0.10$ \\
    BERT \cite{Devlin-NAACL-2019}              & En    &    $79.35\!\pm\!0.08$   & $63.91\!\pm\!0.08$ \\
    BERT \cite{Devlin-NAACL-2019}              & Sp    &    $72.41\!\pm\!0.08$   & $59.87\!\pm\!0.11$ \\
    \hline
    CamemBERT \cite{Martin-ACL-2020}+BERT \cite{Devlin-NAACL-2019}   & Fr+En  &  $\mathbf{82.61}\!\pm\!0.08$   & $\mathbf{65.91}\!\pm\!0.08$ \\
    CamemBERT \cite{Martin-ACL-2020}+BERT \cite{Devlin-NAACL-2019}   & Fr+Sp  &  $81.80\!\pm\!0.08$   & $64.11\!\pm\!0.09$ \\
    CamemBERT \cite{Martin-ACL-2020}+2$\times$BERT \cite{Devlin-NAACL-2019}  & Fr+En+Sp  &  $82.01\!\pm\!0.08$   & $64.95\!\pm\!0.09$ \\
    \hline
  \end{tabular}
  \end{center}
}
\vspace{-0.2cm}
\end{table}

\begin{table}[!th]

  \caption{Results on the development set with distinct fusion techniques applied on three models: two language models trained on French [Fr] and English [En] transcripts, and an audio-based model. We report the mean UAR (in percentages) and the standard deviation over three runs. The best score on each task is highlighted in bold.\vspace{-0.3cm}}
  \label{tab_fusion}
  \small{
  \begin{center}
  \setlength\tabcolsep{3.5pt}
  \begin{tabular}{|l|c|c|c|c|c|}
    \hline
    % Model       & Text [Fr] & Text [En] & Audio & Request UAR & Complaint UAR\\
    \multirow{2}{*}{Model}       & {Text}  & {Text}  & \multirow{2}{*}{Audio} & \multicolumn{2}{c|}{UAR} \\
    \cline{5-6}
    &  [Fr] & [En] & & Request & Complaint\\
    \hline    
    \hline
    Plurality voting      & $\checkmark$ & $\checkmark$ & $\checkmark$  &    $80.08 \pm 0.11$   & $62.11 \pm 0.13$ \\
    \hline
    MLP          & $\checkmark$ &  & $\checkmark$    &    $82.60 \pm 0.07$   & $65.98 \pm 0.07$ \\
    MLP          & $\checkmark$ & $\checkmark$ &     &    $82.61 \pm 0.08$   & $65.91 \pm 0.08$ \\
    MLP          & $\checkmark$ & $\checkmark$ & $\checkmark$    &    $82.65 \pm 0.08$   & $66.08 \pm 0.07$ \\
\hline
    Transformer           & $\checkmark$ &  & $\checkmark$  &    $82.81 \pm 0.08$   & $65.99 \pm 0.09$ \\
    Transformer           & $\checkmark$ &  $\checkmark$  &  &   $82.04 \pm 0.07$   & $65.24 \pm 0.08$ \\
    Transformer           & $\checkmark$ & $\checkmark$ & $\checkmark$  &    $82.81 \pm 0.09$   & $66.13 \pm 0.07$ \\

    \hline
    CCMT (ours)    & $\checkmark$ &    & $\checkmark$ &   $83.01 \pm 0.08$   & $66.20 \pm 0.07$ \\
    CCMT (ours)    & $\checkmark$ &  $\checkmark$  &  &   $81.96 \pm 0.08$   & $65.84 \pm 0.09$ \\
    CCMT (ours)    & $\checkmark$ &  $\checkmark$  & $\checkmark$ &   $\textbf{83.31} \pm 0.08$   & $\textbf{66.64} \pm 0.08$ \\
    \hline
  \end{tabular}
  \end{center}
}
\vspace{-0.4cm}
\end{table}

\noindent
{\bf Results for multiple text modalities.}
The results of the language transformers on three different languages, namely French, English, and Spanish, are presented in Table~\ref{tab_language}. Among the considered language models, the CamemBERT \cite{Martin-ACL-2020} model trained on French data reaches the best performance. This is an expected outcome, since the audio calls are in French, and translating to other languages can introduce translation errors and degrade performance. Still, we believe that fusing models pretrained on different languages can boost the performance of CamemBERT. Therefore, we also explore various combinations between the CamemBERT model and the other models trained on English and Spanish, using an MLP block to fuse the distinct language models. For English and Spanish, we fine-tune BERT \cite{Devlin-NAACL-2019} models that were previously pretrained on corresponding language-specific data. Notably, the BERT model trained on English data outperforms the BERT model trained on Spanish data by approximately $6\%$ in terms of UAR on the request class, and $4\%$ on the complaint class, respectively. Regarding the fusion experiments, the best results are clearly obtained by fusing the French and English models, surpassing the baseline CamemBERT model by approximately $0.3\%$ in terms of UAR for both request and complaint classes. However, the addition of the Spanish BERT model leads to a decrease in performance for both classes. We therefore exclude the Spanish language model from the subsequent experiments.

% \begin{figure*}[!t]
% \begin{center}
% \centerline{\includegraphics[width=0.6\linewidth]{tsne.pdf}}
% %\vspace{-0.25cm}
% \caption{A t-SNE visualization of the CCMT embedding space for the development set. Best viewed in color.}
% \label{fig_tsne}
% %\vspace{-1.1cm}
% %\vspace{-0.9cm}
% \end{center}
% \end{figure*}

\noindent
{\bf Results of multimodal methods.}
In Table \ref{tab_fusion}, we present the results of the multimodal fusion experiments involving three models: CamemBERT \cite{Martin-ACL-2020} trained on French text transcripts, BERT \cite{Devlin-NAACL-2019} trained on English text transcripts, and Wav2Vec2.0 trained on audio samples. 
While fusing the distinct modalities, a consistent pattern emerges across all fusion techniques. Combining CamemBERT with Wav2Vec2.0 \cite{Baevski-NeurIPS-2020} proves to be more effective than combining the two text models, and the most favorable outcomes are consistently achieved when all three modalities are combined. Regarding the fusion techniques, traditional methods such as plurality voting and MLP aggregation demonstrate lower effectiveness compared with more complex approaches based on transformers. When we combine tokens from all modalities into a transformer model, we achieve an UAR of $82.81\%$ for the request class and $66.13\%$ for the complaint class. However, the best results are obtained by combining all modalities via our CCMT model. To this end, we choose CCMT to make our final submissions on the private test set.

\begin{table}[!t]

  \caption{Private test set results of our CCMT model with two or three input modalities, with and without the development set included in the training data. %Regarding the input modalities, \emph{Fr} stands for the CamemBERT model trained on French transcripts, \emph{En} stands for the BERT model trained on English transcripts, and \emph{Audio} stands for the Wav2Vec2.0 model. 
  We alternatively trained the CCMT model on the training set (T), as well as the union between the training and development sets (T+D). The best UAR score on each task is highlighted in bold.\vspace{-0.3cm}}
  \label{tab_private}
  \small{
  \begin{center}
  \begin{tabular}{|c|c|c|c|c|}
    \hline
    Training       & \multirow{2}{*}{Modalities}    & \multicolumn{3}{c|}{UAR} \\
    \cline{3-5}
    data & & Request & Complaint & Average \\
    % Training     & \multirow{2}{*}{Modality} & Request & Complaint & Average\\
    % data & & UAR & UAR & UAR \\
    \hline    
    \hline
    T  & Fr+Audio  &  $85.09\%$   & $64.73\%$ & $74.91\%$ \\
    T  & Fr+En+Audio  &  $\textbf{85.87}\%$   & $\textbf{65.41}\%$ & $\textbf{75.64}\%$ \\
    T+D  & Fr+En+Audio  &  $80.29\%$   & $61.79\%$ & $71.04\%$ \\
    \hline
  \end{tabular}
  \end{center}
}
\vspace{-0.6cm}
\end{table}

% To further assess the discriminative power of CCMT, we illustrate a t-SNE visualization of the feature space derived from the class tokens of our top-scoring CCMT model in Figure \ref{fig_tsne}. The visualization reveals the presence of seven distinct inner clusters, with each cluster containing samples from both classes for both classification tasks, complaint and request. Notably, there is a clear separation within each cluster between samples belonging to distinct request classes. At the same time, it looks like distinguishing samples for the complaint detection task is more challenging due to the less evident separation of the complaint classes. This is consistent with the results reported so far on the two tasks, \ie~requests are generally easier to detect than complaints.

\noindent
{\bf Results on the private test set.}
In Table \ref{tab_private}, we report the results obtained on the private test set using three different approaches. Our first submission is based on a CCMT model that fuses two data modalities via CamemBERT and Wav2Vec2.0 tokens. Our second and third submissions are produced by our full CCMT model, which is based on three modalities. The difference between the second and third submissions lies in the training data, \ie~the second submission uses the official RSC training data, while the third one adds the development set to the training data. 

When using only two data modalities (Fr+Audio), the performance is slightly lower when compared with the performance reached by the complete CCMT model. This observation confirms the beneficial impact of integrating models trained on distinct languages. Including the validation data in the training set appears to degrade our performance, largely due to the challenges involved in selecting a good checkpoint without seeing any validation results. Ultimately, our best submission reaches an UAR of $75.64\%$.

\vspace{-0.1cm}
\section{Conclusion}
\vspace{-0.05cm}

In this paper, we introduced CCMT, a multimodal transformer-based framework designed for request and complaint detection. Our framework incorporates two distinct language models and one audio model, allowing us to effectively capture and analyze information from different modalities. The core component of CCMT is a cascaded cross-attention transformer that iteratively aggregates information from the linguistic and audio features. We evaluated the performance of CCMT in the Requests Sub-Challenge of the ACM Multimedia 2023 Computational Paralinguistics Challenge \cite{Schuller-ACMMM-2023}. Our framework demonstrated outstanding results, achieving an average UAR of $75.64\%$. This performance significantly surpasses the competition baselines (by more than $15\%$), indicating the effectiveness of our approach. %The success of CCMT can be attributed to its ability to leverage the complementary nature of linguistic and acoustic features. By combining information from multiple modalities, our framework captures a more comprehensive representation of the input data, leading to enhanced performance.

{\small
\bibliographystyle{ACM-Reference-Format}
\bibliography{references}
}

%%%%%%%%%%%%%%%%%%%%%%%%%%%%%%%%%%%%%%%%%%%%%%%%%%%%%%%%%%%%

\end{document}